\documentclass[letterpaper]{article} 
\usepackage{aaai24}  
\usepackage{times}  
\usepackage{helvet}  
\usepackage{courier}  
\usepackage[hyphens]{url}  
\usepackage{graphicx} 
\usepackage{natbib}  
\usepackage{caption} 
\usepackage{algorithm}
\usepackage{algorithmic}
\usepackage{amsmath}
\usepackage{multirow}
\usepackage{amsfonts}
\usepackage{newfloat}
\usepackage{listings}
\DeclareCaptionStyle{ruled}{labelfont=normalfont,labelsep=colon,strut=off} 
\floatstyle{ruled}
\newfloat{listing}{tb}{lst}{}
\floatname{listing}{Listing}
\title{Gradual Residuals Alignment: A Dual-Stream Framework for \\GAN Inversion and Image Attribute Editing}
\author{
    Hao Li\textsuperscript{\rm 1}, 
    Mengqi Huang\textsuperscript{\rm 1}, 
    Lei Zhang\textsuperscript{\rm 1}, 
    Bo Hu\textsuperscript{\rm 1}, 
    Yi Liu\textsuperscript{\rm 2}, 
    Zhendong Mao\textsuperscript{\rm 1}\thanks{Corresponding author.}
}
\affiliations{
    \textsuperscript{\rm 1}University of Science and Technology of China, Hefei, China\\
    \textsuperscript{\rm 2}State Key Laboratory of Communication Content Cognition, Beijing, China\\

    \{lihaohn, huangmq\}@mail.ustc.edu.cn, \{leizh23, hubo, zdmao\}@ustc.edu.cn, gavin1332@gmail.com
}

\usepackage{bibentry}

\begin{document}

\maketitle

\begin{abstract}
GAN-based image attribute editing firstly leverages GAN Inversion to project real images into the latent space of GAN and then manipulates corresponding latent codes. Recent inversion methods mainly utilize additional high-bit features to improve image details preservation, as low-bit codes cannot faithfully reconstruct source images, leading to the loss of details. However, during editing, existing works fail to accurately complement the lost details and suffer from poor editability. The main reason is they inject all the lost details indiscriminately at one time, which inherently induces the position and quantity of details to overfit source images, resulting in inconsistent content and artifacts in edited images. This work argues that details should be gradually injected into both the reconstruction and editing process in a multi-stage coarse-to-fine manner for better detail preservation and high editability. Therefore, a novel dual-stream framework is proposed to accurately complement details at each stage. The Reconstruction Stream is employed to embed coarse-to-fine lost details into residual features and then adaptively add them to the GAN generator. In the Editing Stream, residual features are accurately aligned by our Selective Attention mechanism and then injected into the editing process in a multi-stage manner. Extensive experiments have shown the superiority of our framework in both reconstruction accuracy and editing quality compared with existing methods.
\end{abstract}

\section{Introduction}
Image attribute editing, which aims to modify the desired attributes of a given image while preserving other details, has gained increasing research interest for its various real-world applications. Rapid progress has been made in this area with the development of generative adversarial network (GAN) based editing methods, which leverage the latent space of pre-trained GAN models (typically, $\mathcal{W+}$ of StyleGAN) by GAN Inversion \cite{abdal2019image2stylegan}. The critical challenge of GAN Inversion lies in achieving the unity of both high-fidelity details preservation and high editing quality since the distortion-editability trade-off \cite{tov2021designing}.

\begin{figure}[t]
\centering
\includegraphics[width=\columnwidth]{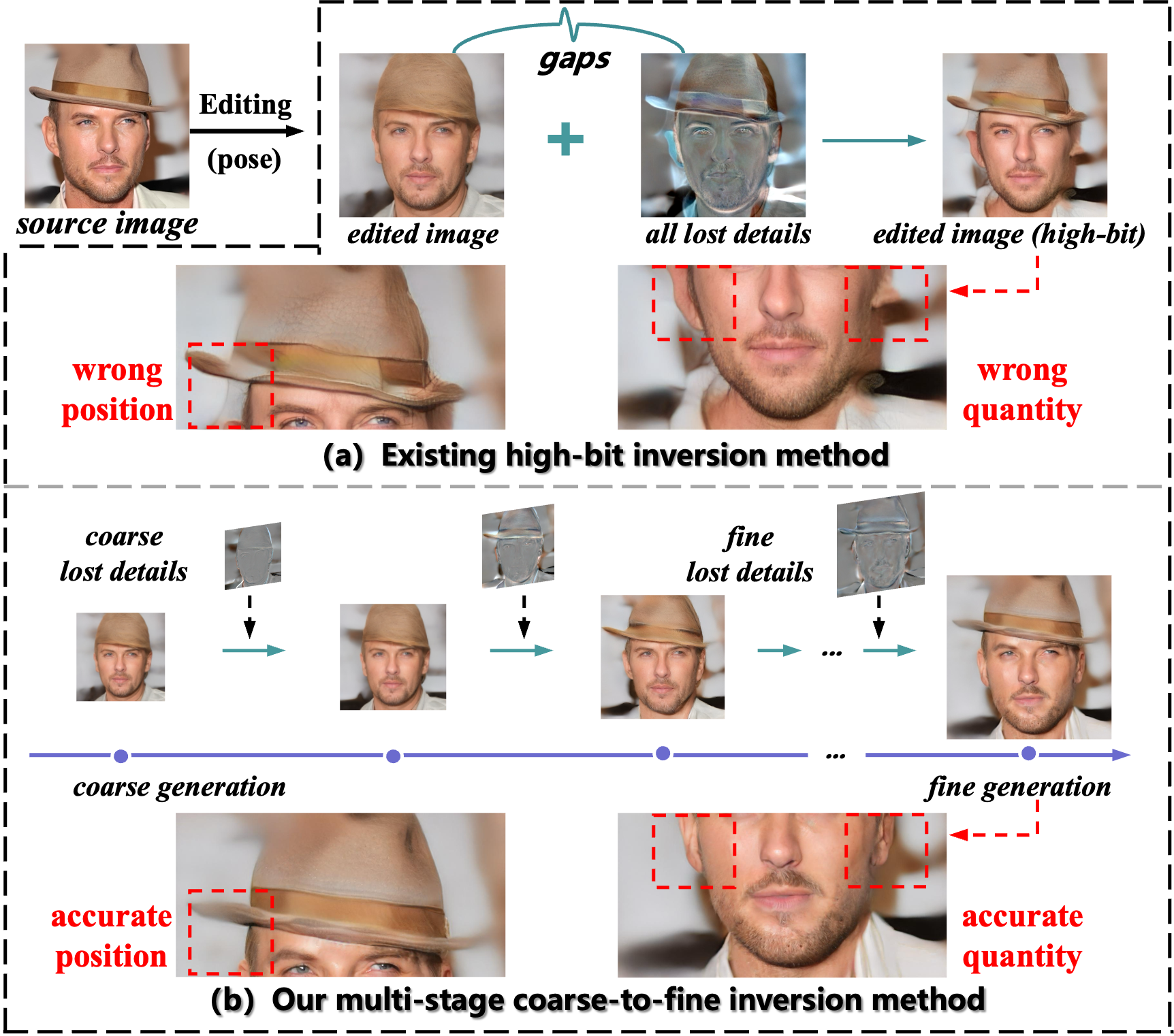} 
\caption{Illustration of our motivation. Giving a source image and then editing (e.g., pose) it. (a) Existing high-bit inversion injects lost details of reconstruction into the edited images as much as possible at one time, which leads to inconsistent content and artifacts. (b) Our method gradually aligns and complements lost details at different stages in editing, which achieves a unity of both high-quality details preservation and high editability with the artifacts mitigated.}
\label{fig1}
\end{figure}

Early GAN Inversion methods \cite{richardson2021encoding,tov2021designing} focus on \emph{Low-bit Inversion} to better map images to low-bit codes (\emph{i.e.}, low-dimension latent codes $\in \mathbb{R}^{18\times512}$), resulting in severe details lost in both of their reconstructed and edited images compared to the source images. Recent works introduced \emph{High-bit Inversion}, which first utilize low-bite codes to generate coarse results and further use high-bit features (\emph{e.g.}, high-dimension feature maps $\in \mathbb{R}^{64 \times 64\times 512}$) to improve details preservation. These high-bit features are derived from source images or the residuals between source images and low-bit codes' reconstruction. For example, \cite{yao2022style,liu2023delving} propose to replace a part of low-bit codes with high-bit features and then restart generation. \cite{wang2022high,pehlivan2023styleres} focus on establishing an additional branch to complement details for image generation by calculating residuals. In conclusion, most recent works pursue injecting source images' high-bit features into the edited images \emph{as much as possible at one time}.

However, existing works overlook the intricate gaps between source images and edited images, and therefore fail to \emph{accurately} complement the lost details for the edited images, leading to poor editability and incoherent generation results. The editing operation itself will bring more or less variations to source images from global layouts to local patterns, and therefore the lost details desired by edited images are also changed synchronously. As shown in Fig.\ref{fig1} (a), the variation of pose leads to differences in details' target position (e.g., the hat) and details' target quantity (e.g., the left ear should have fewer details while the right ear should have more details than the source image). Existing methods inject all the lost details indiscriminately, inherently inducing the position and quantity of details to overfit source images, resulting in inconsistent content and artifacts.

This paper argues that in both reconstruction and editing, lost details should be gradually complemented in a multi-stage coarse-to-fine manner for accurate detail preservation and compelling editability. The reason is that both the reconstruction and editing themselves are, by nature, coarse-to-fine processes with the scale of feature maps increasing, and different-granularity details are required step by step. Gradual addition offers two distinct advantages: (1) Complementing coarse-to-fine details at each step results in cumulative benefits and improves the overall reconstruction quality. (2) The position and quantity of coarse details are easier to align with edited images, which provides a better association foundation for the alignment of finer details, thereby reducing the overall difficulty of the editing. Take Fig.\ref{fig1} (b) as an example. With the coarse-to-fine lost details gradually injected, the content of the edited image becomes closer to the source (\emph{e.g.}, the texture of the hat and ears gradually match the source image), meanwhile, the position and quantity of details will be more accurate (\emph{e.g.}, position of the hat, quantity of ears). This manner can effectively mitigate the issue of artifacts and achieve the unity of both high-fidelity detail preservation and high editing quality.

With this motivation, we propose a novel framework named \textbf{G}radual \textbf{R}esiduals \textbf{A}lignment \textbf{D}ual-Stream Framework for \textbf{Style}GAN inversion and editing (\textbf{GradStyle}), which effectively extracts coarse-to-fine lost details for faithful reconstruction and accurately aligns them with edited images for flexible editing in a multi-stage manner. Specifically, this framework includes a Reconstruction Stream and an Editing Stream, with an Encoding Phase for embedding images. GAN generator blocks are grouped into coarse-to-fine consecutive stages based on their characteristics. In Reconstruction Stream, proposed \emph{Gradual Residual Module} embeds the feature-level distortions between the coarsely reconstructed images and source images into multiple residual features to complement lost details at each stage. A gated fusion mechanism with regularization is further utilized to adaptively fuse residual features in a learnable manner. In the Editing Stream, we propose a novel \emph{Global Alignment Module}, which first achieves an accurate global alignment for residual features based on our \emph{Selective Attention} mechanism, and then adaptively injects them into the editing process. This global alignment provides an effective adjustment for the position and quantity of lost details. To simultaneously train both streams, a self-supervised training strategy without additional labeled edited images is devised.

Our main contributions are summarized as follows:
\begin{itemize}
\item For the first time, we propose a scheme to gradually complement lost details in the reconstruction and editing of images, which achieves a unity of both high-quality detail preservation and high editability in StyleGAN.
\item A novel dual-stream framework, \emph{i.e.,} GradStyle, is proposed to simultaneously conduct reconstruction and editing with a devised self-supervised training strategy. Reconstruction Stream explores coarse-to-fine details information and achieves a more faithful reconstruction, while Editing Stream accurately aligns and adaptively injects these details into the editing process step by step, ensuring better editability. 
\item Extensive experiments have shown the effectiveness of our framework and the improvement over existing methods in terms of both reconstruction accuracy and editing quality, with the generalizability toward various domains.
\end{itemize}

\section{Related Works}

\subsection{GAN Inversion}
GAN Inversion is to invert a given image back into the latent space of a pre-trained GAN model and obtain a latent representation with the capacity to reconstruct it. Recently, the StyleGAN series \cite{karras2019style,karras2020analyzing,karras2020training,karras2021alias} have gained widespread popularity due to its fantastic disentangled latent space which facilitates attribute editing. Our study mainly focuses on StyleGAN inversion.

\noindent \textbf{Low-bit Inversion.} Early optimization-based approaches \cite{zhu2016generative,huh2020transforming,abdal2020image2stylegan++} continuously optimize the latent codes to minimize the reconstruction loss of the source with slow inference. Encoder-based approaches \cite{richardson2021encoding,tov2021designing,hu2022style,mao2022cycle} map latent codes more quickly through a learnable encoder, with better editability but worse fidelity. To keep more details, \cite{wei2022e2style,moon2022interestyle} complement latent codes with the differences between reconstructed and source images. Other works \cite{roich2022pivotal, dinh2022hyperinverter} have attempted to fine-tune the generator. These methods all fail to maintain details due to using low-bit codes only.

\begin{figure*}[t]
\centering
\includegraphics[width=\textwidth]{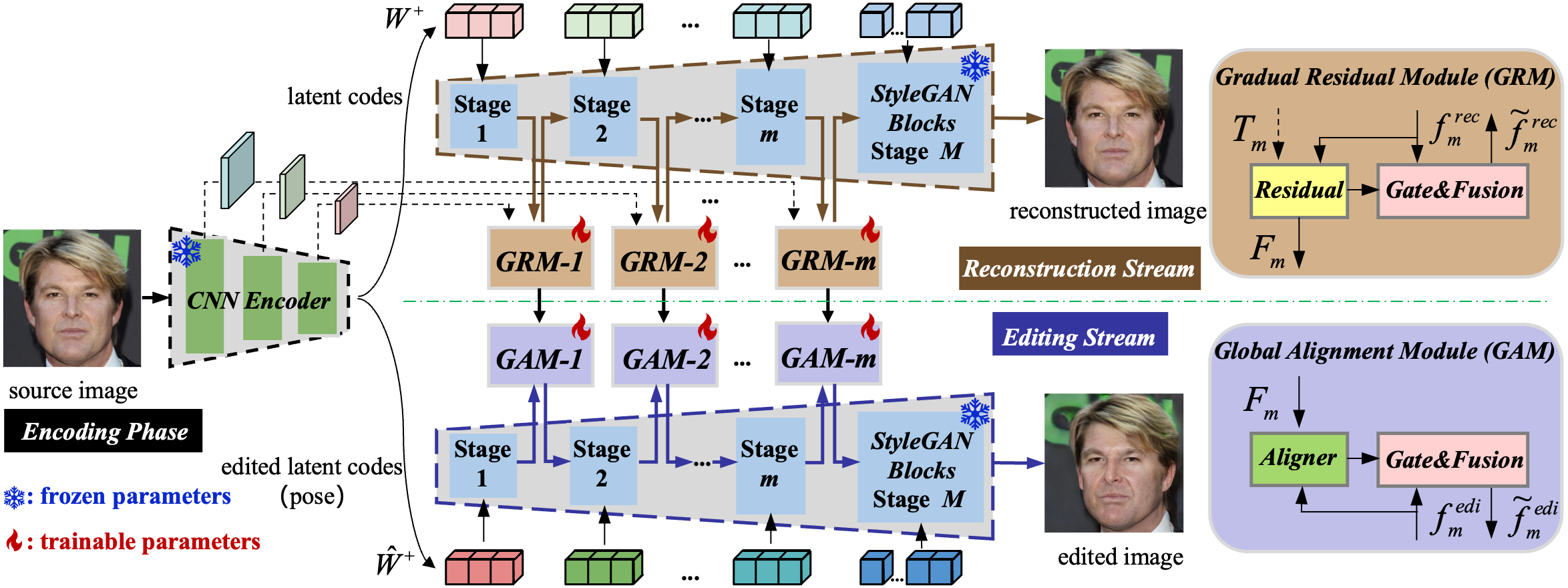} 
\caption{An overview of our dual-stream framework GradStyle. It consists of three parts, an Encoding Phase for embedding images, a Reconstruction Stream for faithful reconstruction and residual features calculation, and an Editing Stream for edited image generation by gradually aligning and adding details information. The proposed Gradual Residual Module and Global Alignment Module are also illustrated, and details of Aligner are especially shown in Fig.\ref{fig3}.}
\label{fig2}
\end{figure*}

\noindent \textbf{High-bit Inversion.} BDInvert \cite{kang2021gan} first proposes using an additional latent space $\mathcal{F}$. HFGI \cite{wang2022high} utilizes the image-level distortions between the source and the reconstructed, but image-level features retain excessive spatial dependencies. Other methods \cite{yao2022style,liu2023delving} train an encoder to obtain low-bit codes and high-bit features simultaneously, then replace the first several latent codes with high-bit features. StyleRes \cite{pehlivan2023styleres} actually packs all details information into a single residual feature and adds it at one stage. The above methods suffer from severe artifacts, however, our approach, which employs a multi-stage manner to gradually complement coarse-to-fine details, can effectively suppress artifacts.

\subsection{Latent Space Editing}
Manipulating latent codes in the latent space focuses on searching meaningful editing directions for interpolation. For supervised methods, off-the-shelf attribute classifiers are employed to obtain attribute labels and then analyze the spatial distribution of latent codes for editing directions. For example, InterfaceGAN \cite{shen2020interfacegan} utilizes Support Vector Machines (SVMs) to learn a classification hyperplane. \cite{abdal2021styleflow, wang2021hijack} employs neural networks to distinguish directions between different attributes. For unsupervised methods, GANspace \cite{harkonen2020ganspace} applies Principal Component Analysis (PCA), \cite{shen2021closed} decomposes the model weights of GAN networks. Moreover, with the advancement of multimodal techniques \cite{radford2021learning}, language-based methods \cite{wu2021stylespace, patashnik2021styleclip} have further expanded the application area.

\section{Methodology}
The overall framework is depicted in Fig.\ref{fig2}. In the Encoding Phase (section 3.1), an encoder is adopted to embed source images to both low-bit latent codes and hierarchical features. Reconstruction Stream (section 3.2) employs \emph{Gradual Residual Module (GRM)} to calculate residual features and faithfully reconstruct images. In Editing Stream (section 3.3), we utilize \emph{Global Alignment Module (GAM)} to align residual features with edited images and inject them into the generator step by step. Finally, a self-supervised training strategy (section 3.4) is conducted to train two streams simultaneously. Next, we will describe them in detail.

\noindent\textbf{Notations.} Formally, $X\in \mathbb{R}^{H_0\times W_0\times 3}$ denotes source images. Both reconstruction and editing utilize the same generator from pre-trained StyleGAN but are driven by different latent codes. $N$ is the number of latent codes of each image, $M$ is the total number of generation stages, $T_m$ and $F_m$ respectively denote the hierarchical features from the encoder and the residual features from \emph{GRM} at stage $m$. $X^{r}$ and $X^{e}$ is reconstructed and edited images, $f_m^{rec}$ and $f_m^{edi}$ denote the corresponding feature maps of generator blocks at stage $m$.

\subsection{Encoding Phase}
With the pre-trained CNN encoder $E_0$ from \cite{tov2021designing}, we have $T, W^{+}=E_0 (X)$, where $W^{+}=\{{w}_i|i=1, 2, \cdots, N, \; {w}_i \in \mathbb{R}^{512}\}$ are latent codes and can coarsely reconstruct $X$. Our encoder is based on a pyramid structure to generate latent codes, corresponding hierarchical features $T = \{T_m|m=1,2,\cdots,M\}$ can be naturally obtained from the different layers of the hierarchical encoder, where $T_m \in \mathbb{R}^{H_m\times W_m\times c }$, $(H_m,W_m)=(H_0/n_m,W_0/n_m), c=512$. These hierarchical features represent the coarse-to-fine details information of the source images. Further, for our editing stream, as $W^{+}$ can be interpolated by meaningful direction \cite{shen2020interpreting}, we obtain edited latent codes $\hat{W}^{+}=W^{+}+\alpha \Delta W^{+}$, where $\Delta W^{+}$ is the editing direction and $\alpha$ is the editing amplitude.

\subsection{Reconstruction Stream}
This stream targets a faithful reconstruction with the input of $W^{+}$ and $T$, and calculating residual features $F_m$ for the Editing Stream. Each ${w}_i$ controls a StyleGAN block, i.e., Modulated Convolution layer \cite{karras2020analyzing}, and the different block affects different content from coarse (e.g., shapes of face) to fine levels (e.g., wrinkles). Instead of naively refining all blocks, we propose to selectively refine several key blocks (which are enough to complement all lost details) to further improve efficiency. Specifically, as shown in Fig.\ref{fig2}, all generator blocks with corresponding latent codes are grouped into coarse-to-fine consecutive $M$ parts, and each part is treated as a generation stage. We then insert a \emph{GRM} between every two stages. At stage m, with hierarchical features $T_m$ and the output of generator block $f_m^{rec}$, we calculate that:
\begin{equation}
  \tilde{f}_m^{rec}, F_m = {GRM}_m(T_m,f_m^{rec}),
  \label{eq:GRM_Block}
\end{equation}
where $F_m$, $f_m^{rec}$ and $\tilde{f}_m^{rec}\in \mathbb{R}^{H_m\times W_m\times c }$, and $\tilde{f}_m^{rec}$ includes richer details than $f_m^{rec}$, thereby serving as the input feature of the next stage. After crossing all stages, we can obtain a well-reconstructed image.

\textbf{Gradual Residual Module (GRM) and Gate\&Fusion.} It is essential to find out what details the current stage fails to reconstruct concerning the source image, and then complement them, so we design \emph{Gradual Residual Module}. In each \emph{GRM}, we utilize the ResNet-based network $E_{res}$ to obtain the residual features between $T_m$ and $f_m^{rec}$, 
\begin{equation}
  F_m=E_{res}(W_{T}T_m,W_{f}f_m^{rec}),
  \label{eq:res_Block}
\end{equation}
for simplicity, we employ $W_{T}$ and $W_{f}$ to denote learnable convolution networks, which transfer both features to the same semantic space. Further, we utilize the \emph{Gate\&Fusion} to learn how to adaptively fuse the residual features, as not all details are required to be added at this stage. We need to choose them in a gating manner according to the characteristics of different stages:
\begin{equation}
  g_m =\sigma(W_{g} \left[ f_m^{rec},F_m\right ]),
  \label{eq:gate_Block}
\end{equation}
\begin{equation}
  \tilde{f}_m^{rec} = f_m^{rec} + g_m\cdot F_m,
  \label{eq:fusion_Block}
\end{equation}
where $\sigma(\cdot)$ is a \emph{sigmoid} function, $W_{g}$ is the learnable layers. The gating maps $g_m$ share the same size with $F_m$ and can determine which patches of $F_m$ are used at this stage. Residual features can be adaptively selected by \emph{Gate\&Fusion}, so an extra $L_1 \; Regularization$ term is utilized to avoid the overfitting of details stemming from redundant information:
\begin{equation}
  \mathcal{L}_{f}=\sum_{m=1}^{M} \left \| \tilde{f}_m^{rec}-f_m^{rec} \right \|_1 = \sum_{m=1}^{M} \left \| g_m\cdot F_m \right \|_1.
  \label{eq:loss_5}
\end{equation}

\subsection{Editing Stream}
Generator blocks of this stream are grouped in the same way as the Reconstruction Stream, and each \emph{GAM} is also inserted between two stages. With the edited latent codes $\hat{W}^{+}$, each stage outputs an edited feature map $f_m^{edi}$. Our \emph{GAM} receives the residual features $F_m$ from \emph{GRM} at each stage and then aligns $F_m$ with $f_m^{edi}$ through \emph{Aligner} block, finally getting $\tilde{f}_m^{edi}$ which will include aligned lost details. That is
\begin{equation}
  \tilde{f}_m^{edi} = {GAM}_m(F_m,f_m^{edi}).
  \label{eq:GAM_Block}
\end{equation}
After the aligned details information is added across the whole generation process, the edited image with more accurate and consistent details can be obtained.

\begin{figure}[t]
\centering
\includegraphics[width=0.98\columnwidth]{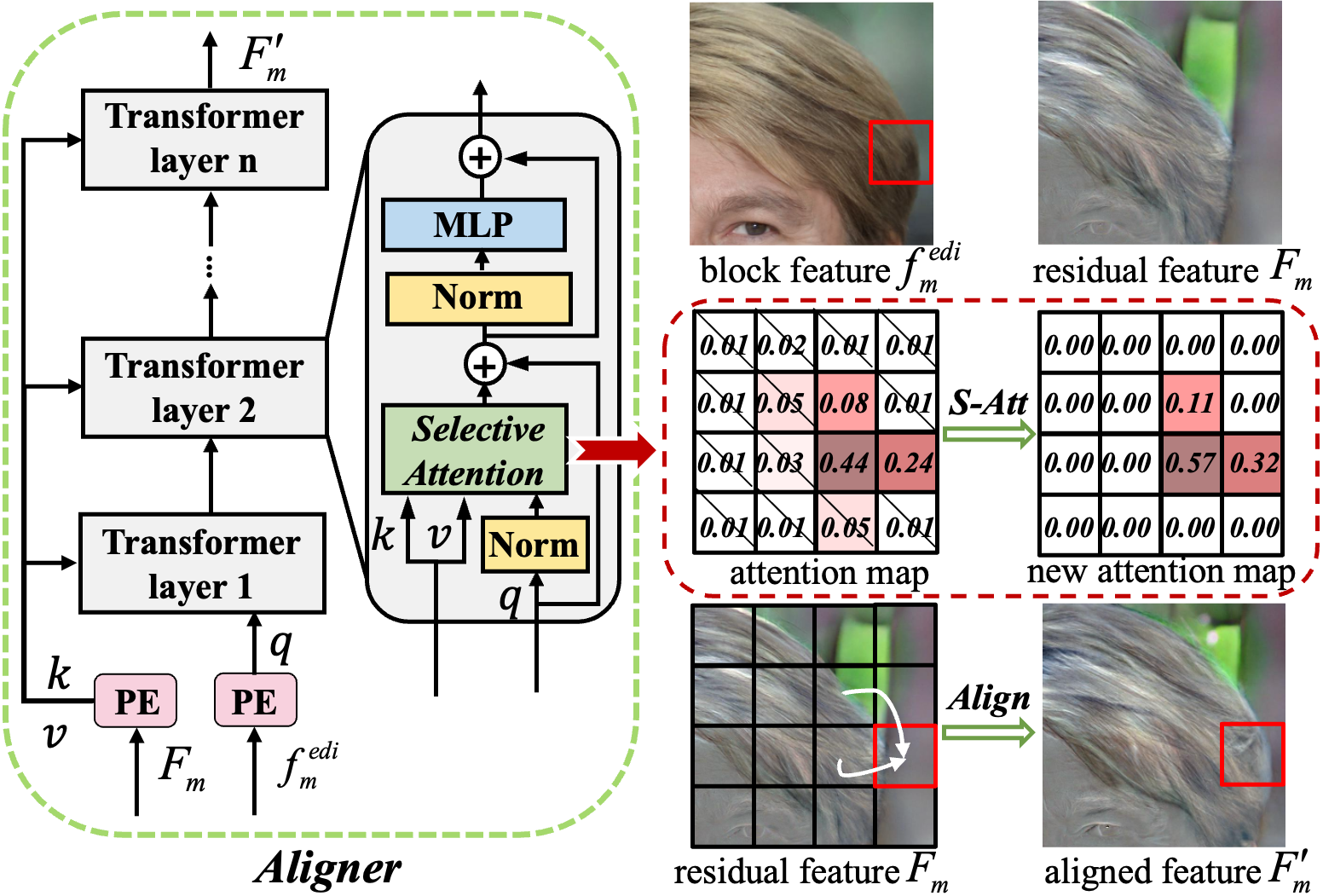} 
\caption{Detailed structure of Aligner block and an image-level visualization example for Selective Attention (we actually utilize it in the feature level). In the 1st row of the example, a coarsely edited image stands for the block feature $f_m^{edi}$ (query), and the unaligned residual feature $F_m$ (key and value) is on its right. In the 2nd row, for a region of query, its attention map indicates that there are many irrelevant regions, Selective Attention will suppress irrelevant regions and enhance relevant regions. The last row shows that a region of $F_m^{'}$ is combined by similar regions of unaligned $F_m$.}
\label{fig3}
\end{figure}

\textbf{Global Alignment Module (GAM) and Selective Attention.} \emph{GAM} is aimed to achieve a more accurate global alignment according to the semantic correlation between unaligned residual features $F_m$ and edited feature map $f_m^{edi}$. The main basis is that $F_m$ inherently inherits the semantic characteristics of reconstructed image feature map $f_m^{rec}$, which has a strong content-level correlation with $f_m^{edi}$. So we design a novel attention-based \emph{Aligner} block for this alignment, as shown in Fig.\ref{fig3}. Thanks to the Transformer structure, we can deal with various editing scenes no matter how the position and quantity of details change, as this structure owns the long-range awareness to associate and fuse similar region features. \emph{Aligner} is represented as
\begin{equation}
  {F}_m^{'} = E_{ab}(F_m,f^{edi}_m),
  \label{eq:atten_0}
\end{equation}
where ${F}_m^{'}$ is the aligned residual features and $E_{ab}$ is the \emph{Aligner} block at each stage. 

In details, the input feature $F_m$ and $f^{edi}_m\in\mathbb{R}^{H_m\times W_m\times c}$ are first flattened to $\mathbb{R}^{L\times c}$, where $L=H_m\times W_m$, and then both go through a Sinusoidal Positional Embedding layer \emph{PE} \cite{vaswani2017attention}, getting $Z_F,Z_{f}$. 
The key component for alignment is our \emph{Selective Attention} mechanism and we apply it several times for fully exploring semantic correlation. It can be mathematically formed as:
\begin{equation}
  q,k,v = W_q Z_{f},W_k Z_F,W_v Z_F,
  \label{eq:atten_1}
\end{equation}
\begin{equation}
S\mbox{-}Att(Z_F,Z_{f})=\mathrm{Softmax}(\frac{qk^{T}\odot \mathrm{Top}_{\mu}(qk^{T})}{\sqrt{d_k}})v,
  \label{eq:atten_2}
\end{equation}
where $W_q$, $W_k$, $W_v$ are the learnable parameters, scaling factor $d_k=64$, and the multi-head mechanism is employed. Eq.\ref{eq:atten_2} indicates that only the top $\mu\% $ values of $qk^{T}$ will undergo the $\mathrm{Softmax}$ operation to calculate the attention map, while the remaining values will be suppressed, as shown in Fig.\ref{fig3}. Based on our \emph{Selective Attention}, \emph{Aligner} can not only align $F_m$ by combining the relevant regions but also weaken the influence of irrelevant regions. Following \emph{GRM}, our \emph{GAM} also employ \emph{Gate\&Fusion} block to adaptively fuse features, and finally get $\tilde{f}_m^{edi}$.

\subsection{Training}
\textbf{Self-supervised Training.} During training, the encoder and StyleGAN2 generator blocks are all fixed, the key is how to seamlessly train our \emph{GRMs} and \emph{GAMs}. All \emph{GRMs} will be updated under the guidance of the reconstruction error of the source $X$, with the gradient from \emph{GAMs} detached. For the Editing Stream, alleviating the misalignment between residual features and edited feature maps is the ultimate goal of \emph{GAMs}. The training of \emph{GAMs} necessitates enough misaligned feature pairs, but the absence of manually annotated edited images precludes the availability of ground truth. 

A self-supervised training strategy is devised for our Editing Stream. We set $\alpha=0$ in $\hat{W}^{+}=W^{+}+\alpha \Delta W^{+}$ in the Encoding Pharse, which implies Editing Stream will generate the same output images as the Reconstruction Stream and can also calculate loss based on the source $X$. As the same latent codes result in the same generator block features, received residual features $F_m$ will become well-aligned with the edited image feature maps $f^{edi}_m$. To train their alignment ability, we augment $F_m$ with random perspective transformation \cite{wang2022high} to simulate the layout misalignment with $f^{edi}_m$, that is $\hat{F}_m = \mathrm{Trans}(F_m)$, thereby getting misaligned feature pairs $\{\hat{F}_m,f^{edi}_m\}$ with the ground truth $F_m$. The \emph{Aligner} block is encouraged to produce aligned residual features $F^{'}_m = E_{ab}(\hat{F}_m,f^{edi}_m)$. We take $F_m$ as the intermediate supervision signal and utilize an aligner loss:
\begin{equation}
  \mathcal{L}_{aligner}=\sum_{m=1}^{M} \left \| F^{'}_m-F_m \right \|_1.
  \label{eq:self-super}
\end{equation}
For the consistency of feature discrimination abilities, \emph{Gate\&Fusion} block in \emph{GAMs} share weights with \emph{GRMs'}. 

\noindent \textbf{Training Losses.} The source is $X$, and the output of two streams are $X^{r}$ and $X^{e}$. Following \cite{tov2021designing}, we first employ $L_2$ loss, $lpips$ loss for faithful reconstruction:
\begin{equation}
  \mathcal{L}_{l2}= \left \| X^{r}-X \right \|_2 +  \left \| X^{e}-X \right \|_2,
  \label{eq:loss_1}
\end{equation}
\begin{equation}
  \mathcal{L}_{lpips}= \left \| \Phi(X^{r})-\Phi(X) \right \|_2 +  \left \| \Phi(X^{e})-\Phi(X) \right \|_2,
  \label{eq:loss_2}
\end{equation}
where $\Phi(\cdot)$ is the pre-trained VGG network \cite{simonyan2014very}. ID loss is to keep the identity consistent,
\begin{equation}
  \mathcal{L}_{id}=(1-\left \langle F(X),F(X^{r}) \right \rangle )+(1-\left \langle F(X),F(X^{e}) \right \rangle ),
  \label{eq:loss_3}
\end{equation}
where $F(\cdot)$ is pre-trained ArcFace \cite{deng2019arcface} or a ResNet for different domains \cite{tov2021designing}. 

For better image quality, we also utilize an adversarial loss $\mathcal{L}_{adv}$ based on a discriminator $D$, Which is initialized with well-trained parameters from StyleGAN2 and then trains along with our framework.
\begin{equation}
  \mathcal{L}_{adv}= -\mathbb{E}[log(D(X^{r}))]-\mathbb{E}[log(D(X^{e}))].
  \label{eq:loss_4}
\end{equation}

The overall loss is a weighted sum of the above losses:
\begin{equation}
\begin{split}
    \mathcal{L}= \mathcal{L}_{l2}+\lambda_{lpips}\mathcal{L}_{lpips}+\lambda_{id}\mathcal{L}_{id}+\lambda_{adv}\mathcal{L}_{adv}\\+\lambda_{f}\mathcal{L}_{f}+\lambda_{aligner}\mathcal{L}_{aligner}.
  \label{eq:loss_6}
\end{split}
\end{equation}
See the Appendix for hyperparameters and more details. After training, our framework can inverse images through the Reconstruction Stream and meanwhile conduct attribute editing in the Editing Stream.  
\section{Experiments}
\subsection{Settings}
\textbf{Experimental setup.} Our approach is based on pre-trained StyleGAN2 \cite{karras2020analyzing} and e4e encoder \cite{tov2021designing}. Main experiments are conducted in the face-domain dataset, we use the FFHQ \cite{karras2019style} to train and the Celeba-HQ \cite{karras2017progressive} to evaluate. During image editing, we choose off-the-shelf InterfaceGAN \cite{shen2020interfacegan} and GANspace \cite{harkonen2020ganspace} as latent code editors. For generalizability evaluation, we also test our method in the different domain datasets, including Stanford Car \cite{krause20133d} for car and Metface \cite{karras2020training} for artistic portrait.

\noindent \textbf{Implementation details.} Our framework is mainly trained on face-domain images with $1024\times1024$ resolution ($N=18$ latent codes in total), adopting Adam optimizer \cite{kingma2014adam} with LookAhead technique \cite{zhang2019lookahead}. In all experiments, both streams of our framework are 4-stage processes, which means the 3rd, 6th, and 8th blocks of the StyleGAN generator are refined. Other details are included in our Appendix.

\begin{figure*}[t]
\centering
\includegraphics[width=\textwidth]{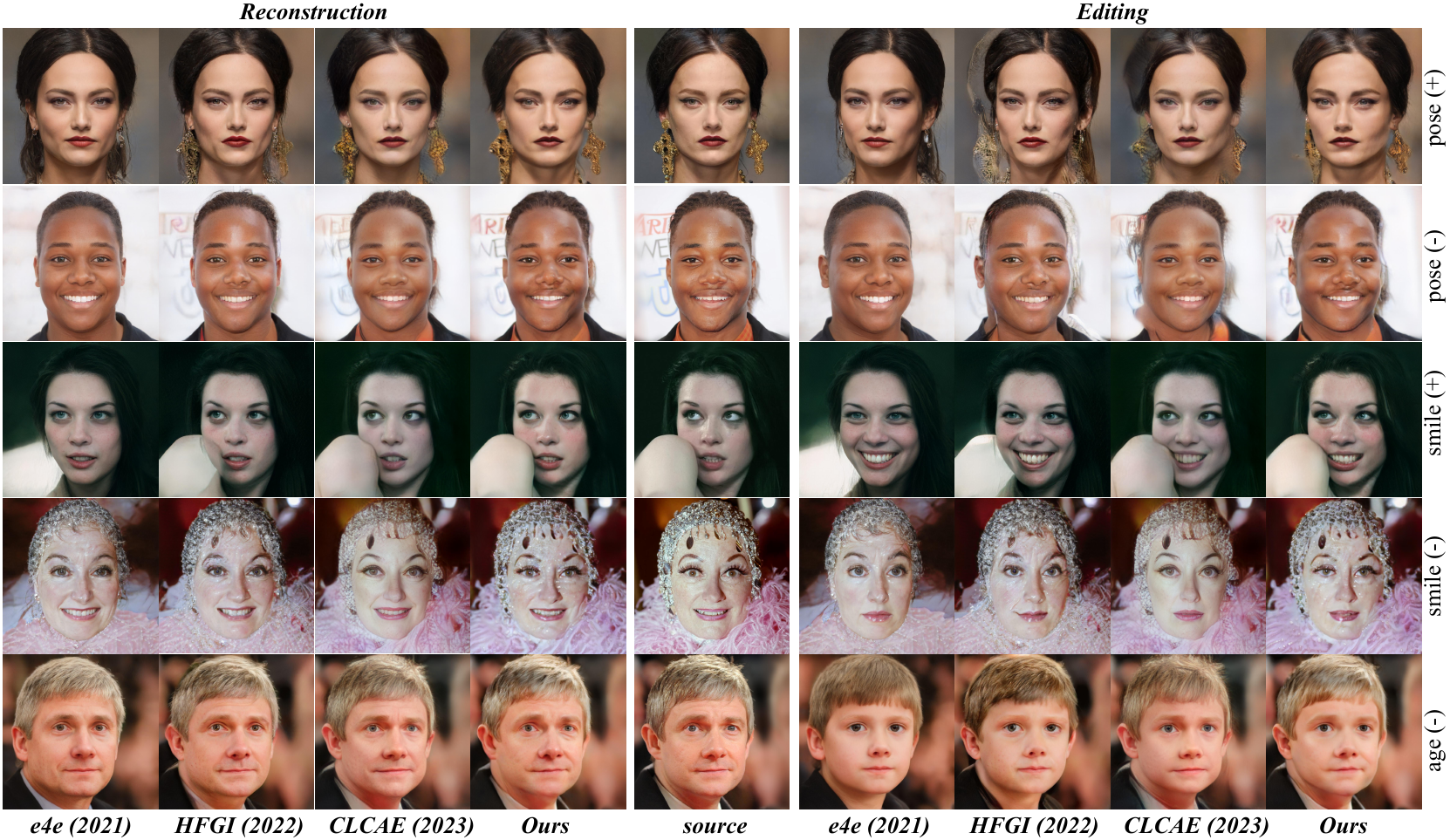} 
\caption{Qualitative results of reconstruction and editing. The left shows reconstructed results from several recent methods and our method, and the right shows edited results based on InterfaceGAN \cite{shen2020interfacegan}. The source is in the middle.}
\label{fig4}
\end{figure*}

\subsection{Evaluations}
\textbf{Quantitative results.} To verify the reconstruction fidelity and editability, we compare our method quantitatively with other state-of-the-art methods. Low-bit Inversion includes e4e \cite{tov2021designing}, ReStyle \cite{alaluf2021restyle} and HyperStyle \cite{alaluf2022hyperstyle}, for High-bit Inversion, HFGI \cite{wang2022high}, StyleRes \cite{pehlivan2023styleres} and CLCAE \cite{liu2023delving} have been included. We compare them by reporting some metrics that are calculated on the highest resolution of the first 1000 images from Celeba-HQ. We adopt L2 distance, LPIPS \cite{zhang2018unreasonable}, and SSIM \cite{wang2004image} to measure pixel-level, feature-level, and structure-level similarity between source and reconstructed images. A pre-trained identity-recognition network \cite{huang2020curricularface} is employed to measure identity similarity (ID), and we also report the Peak Signal-to-Noise Ratio (PSNR).

As shown in our Table.\ref{table1}, we report quantitative comparisons of reconstruction quality. High-bit Inversion has better results than Low-bit Inversion on almost all metrics. Our method achieves the best results among all competing methods on all metrics, implying our coarse-to-fine strategy helps to generate richer and more accurate image content. Most significantly, we have achieved a significant improvement in identity preservation, which means our method has a better ability to keep identity details. 

In Table.\ref{table2}, we show quantitative comparisons of attribute editing. As the smile always involves the quantity change of details while the pose involves the position change, we choose them as representatives. We add or remove the smile and pose in our test images, and then calculate the average ID score as the straight quantitative measurement to evaluate editing performance since other metrics are no longer suitable for editing. It shows our method works better in attribute control, implying that we have flexibly manipulated the special attribute with enough details preserved.

\setlength{\tabcolsep}{2.7pt}
\begin{table}[t]
\small
\begin{center}
\renewcommand{\arraystretch}{1.02}
\begin{tabular}{l|ccccc}
\hline
\multicolumn{1}{l}{Method} & \multicolumn{1}{c}{ID($\uparrow$)} & \multicolumn{1}{c}{SSIM($\uparrow$)} & \multicolumn{1}{c}{L2($\downarrow$)} & \multicolumn{1}{c}{LPIPS($\downarrow$)} & \multicolumn{1}{c}{PSNR($\uparrow$)} \\ \hline
    e4e   & 0.499 & 0.605 & 0.053 & 0.394 & 19.124 \\
    ReStyle   & 0.506 & 0.607 & 0.049 & 0.384 & 19.462 \\
    HyperStyle   & 0.697 & 0.627 & 0.035 & 0.352 & 21.023 \\ \hline
    HFGI   & 0.682 & 0.645 & 0.027 & 0.328 & 22.065 \\
    StyleRes   & 0.758 & 0.674 & 0.019 & 0.286 & 23.603 \\
    CLCAE   & 0.719 & 0.687 & 0.016 & 0.289 & 24.375 \\ \hline
    \textbf{GradStyle (ours)}   & \textbf{0.813} & \textbf{0.696} & \textbf{0.015} & \textbf{0.269} & \textbf{24.583}\\ \hline
\end{tabular}
\caption{Quantitative comparisons of reconstruction quality, $\downarrow$ indicates lower is better while $\uparrow$ indicates higher is better.}
\label{table1}
\end{center}
\end{table}

\setlength{\tabcolsep}{3pt}
\begin{table}[t]
\small
\begin{center}
\renewcommand{\arraystretch}{1.02}
\begin{tabular}{l|cccc}
\hline
\multirow{2}{*}{Method} & \multicolumn{1}{c}{Pose(+)} & \multicolumn{1}{c}{Pose(-)} & \multicolumn{1}{c}{Smile(+)} & \multicolumn{1}{c}{Smile(-)} \\ \cline{2-5}
          & \multicolumn{4}{c}{ID($\uparrow$)} \\ \hline
    e4e   & 0.464 & 0.461 & 0.446 & 0.379 \\
    ReStyle   & 0.487 & 0.487 & 0.468 & 0.428 \\
    HyperStyle   & 0.641 & 0.651 & 0.608 & 0.577 \\ \hline
    HFGI   & 0.556 & 0.541 & 0.544 & 0.480 \\
    StyleRes   & 0.581 & 0.584 & 0.583 & 0.556 \\
    CLCAE   & 0.675 & 0.672 & 0.653 & 0.637 \\ \hline
    \textbf{GradStyle (ours)}   & \textbf{0.677} & \textbf{0.689} & \textbf{0.690} & \textbf{0.671} \\ \hline
\end{tabular}
\caption{Quantitative comparisons of attribute editing. (+) stands for adding this attribute while (-) stands for removing.}
\label{table2}
\end{center}
\end{table}

\noindent \textbf{Qualitative results.} To visually demonstrate the advantages of our method, we have compared it with other three recent representative methods in Fig.\ref{fig4}: e4e \cite{tov2021designing}, HFGI \cite{wang2022high} and CLCAE \cite{liu2023delving}. We add or remove three types of facial attributes: pose, smile, and age in human faces, and show their reconstructed images on the left of Fig.\ref{fig4} while the edited images are on the right. 

We modify the pose of faces in the first two rows of Fig.\ref{fig4}. We can see that e4e easily edits images but loses many details, such as earrings (1st row) and the background (2nd row). HFGI suffers from reconstruction errors and severe silhouette \emph{artifacts} in editing. CLCAE can keep the most details in reconstruction but it fails to flexibly edit (1st and 2nd row). Better than all, our method can correctly align lost details without \emph{artifacts}. In 3rd row, our method generates more natural teeth with the arm preserved but other methods fail to do both. For other rows, we can observe that our method preserves more details in both reconstruction and editing, such as more faithful clothes and headwear (4th row), and the more similar hairstyle and face (5th row). In short, our method achieves the most faithful reconstruction and the best editing quality among these methods.

\begin{figure}[t]
\centering
\includegraphics[width=\columnwidth]{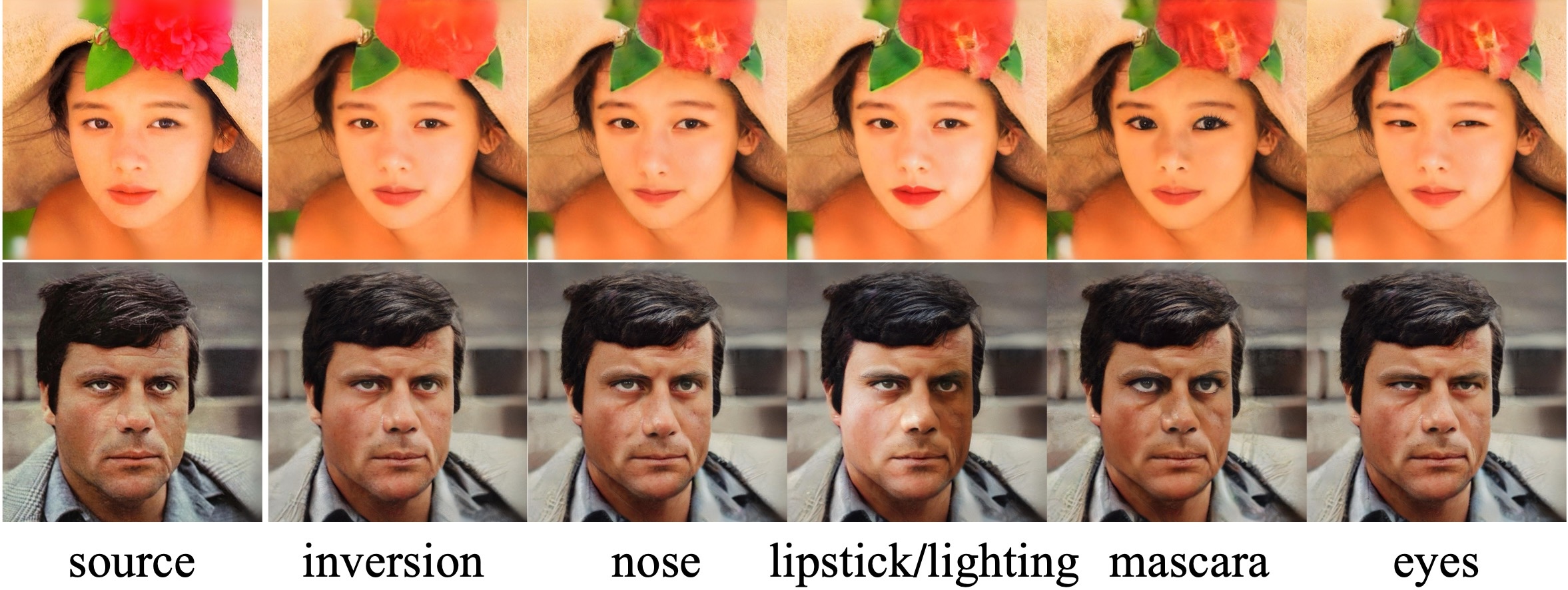} 
\caption{Generalizability of our self-supervised training strategy to deal with various attributes.}
\label{fig5}
\end{figure}

\begin{figure}[t]
\centering
\includegraphics[width=\columnwidth]{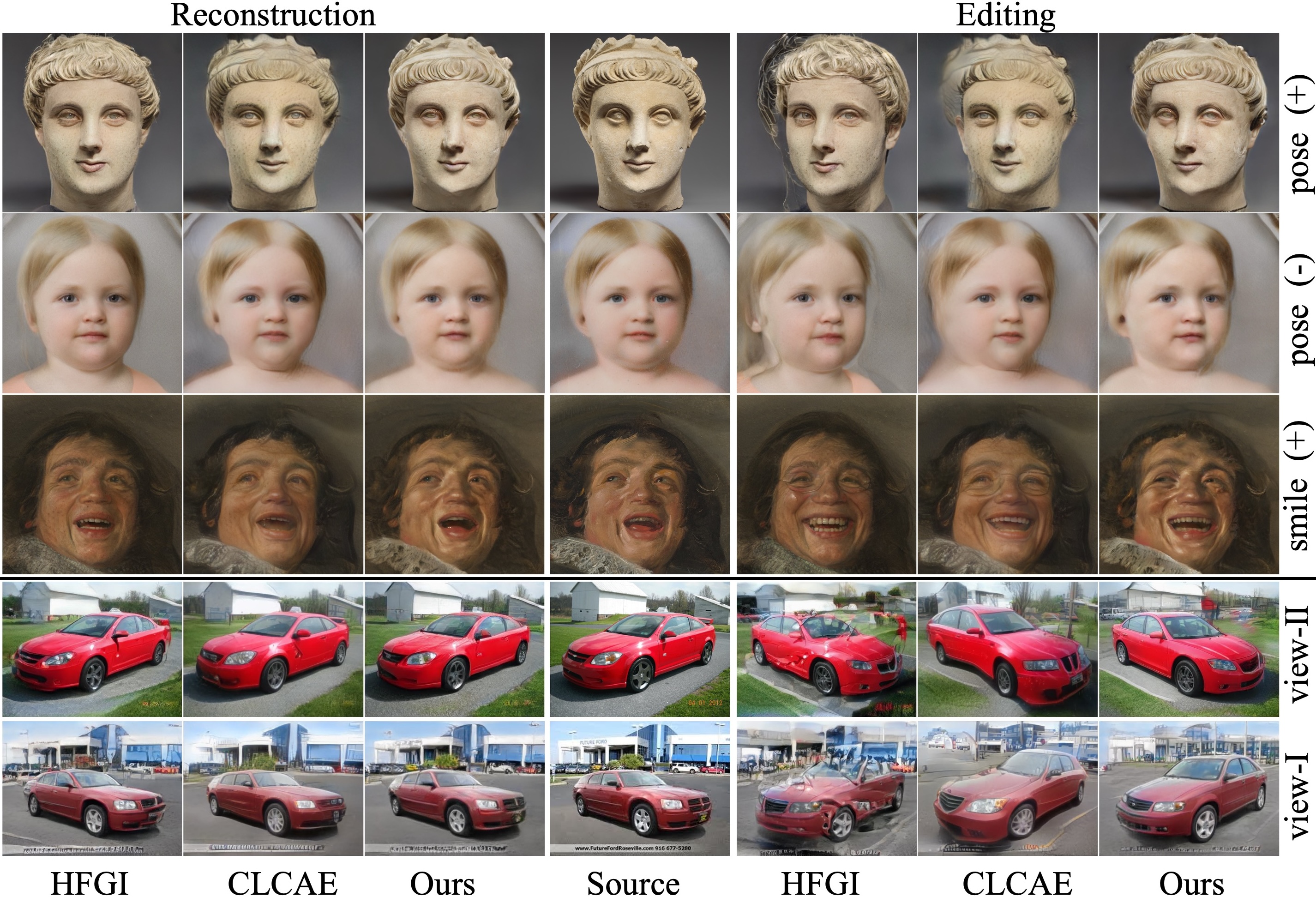} 
\caption{Generalizability of our method in different domains with reconstruction (left) and editing (right).}
\label{fig6}
\end{figure}

\noindent \textbf{Generalizability.} In this work, we use a self-supervised training strategy to train our framework without any labeled edited images. However, what inspires us is that the performance of our framework in the manipulations of various attributes (such as nose, lipstick, lighting, mascara, and eyes) is also plausible, as shown in Fig.\ref{fig5}. These editings are based on InterfaceGAN \cite{shen2020interfacegan}. It implies that the self-supervised training strategy can train our framework's universal editing capability regardless of the editing scenes. 

To evaluate the generalizability of our method in different domains, we further illustrate the results in Fig.\ref{fig6}, comparing with HFGI and CLCAE. For artistic portraits, we train our framework in the FFHQ dataset (i.e., human face domain) and only test on those out-of-domain images without any fine-tuning. We can see that our method works better than all other methods in both reconstruction and editing. For cars, we both train and test in the car domain, more details have been included in Appendix. We illustrate two difficult editing scenes in the last two rows of Fig.\ref{fig6}, and it shows our method can generate more realistic cars and keep closer background details than other methods. All of these evaluations have demonstrated our framework can work well in various domains without overfitting into a specific domain.

\begin{figure}[t]
\centering
\includegraphics[width=\columnwidth]{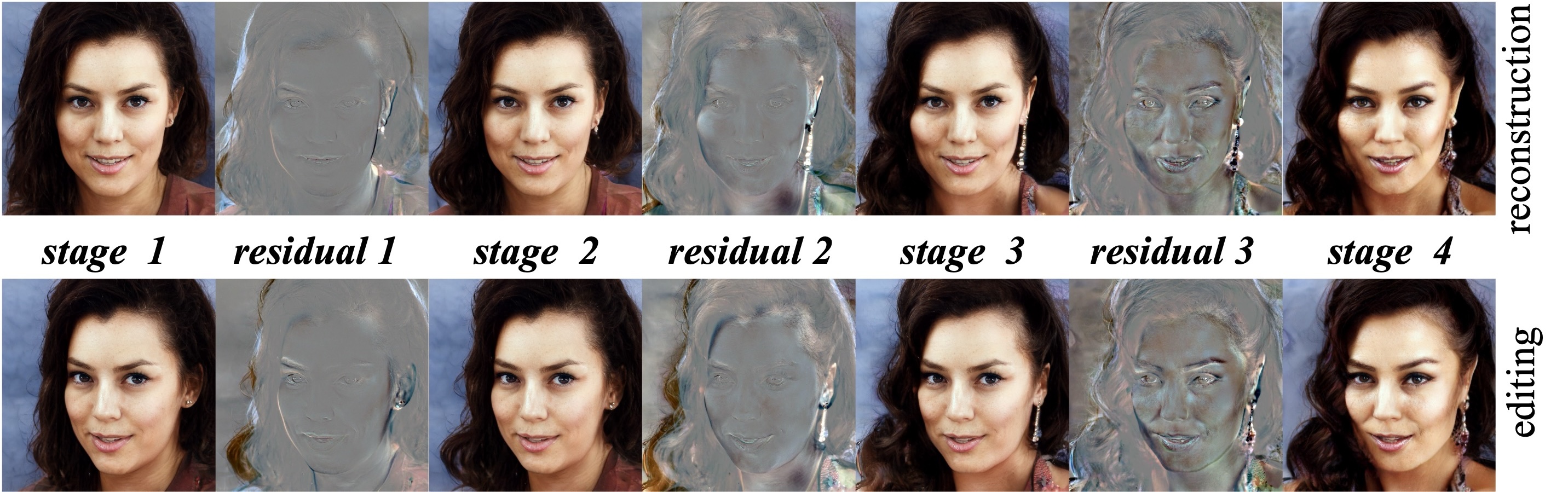} 
\caption{Visualization of our generation results and residuals in different stages of reconstruction and editing.}
\label{fig7}
\end{figure}

\begin{figure}[t]
\centering
\includegraphics[width=\columnwidth]{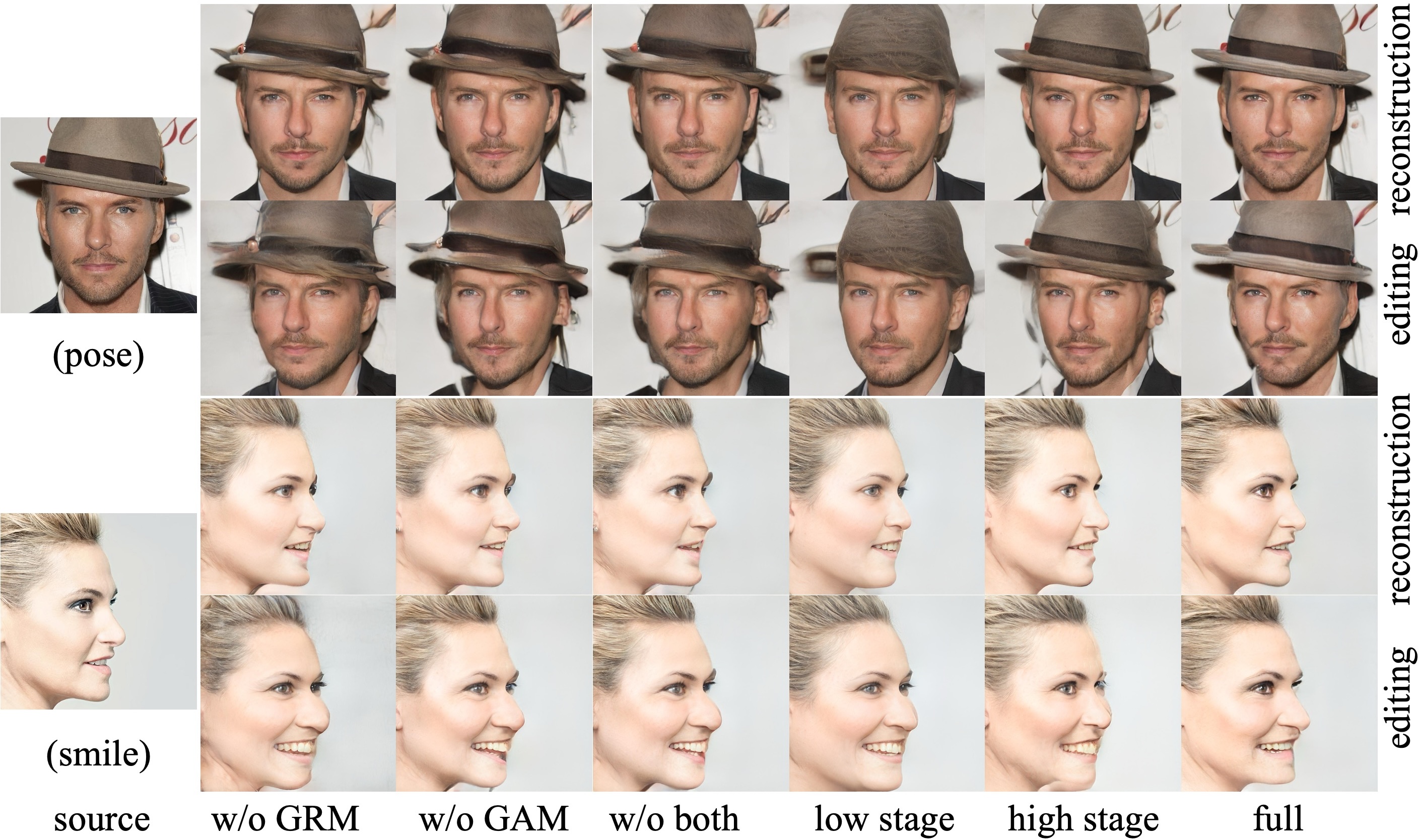} 
\caption{Ablation studies, where low (high) stage indicates only adding residuals in a single low-level (high-level) stage.}
\label{fig8}
\end{figure}

\noindent \textbf{Visualization of residual additions.} Our method employs a coarse-to-fine manner to add residual features. A visualization example is illustrated in Fig.\ref{fig7}. We can notice that the residual features between stage 1 and stage 2 mainly include coarse details, such as hair and clothes shape. With the development of stages, details become finer. Our framework can effectively align each residual feature with the edited layout, resulting in high-quality editing.

\noindent \textbf{Ablation study.} The different effects of each proposed component are compared in Fig.\ref{fig8}. By showing the change in the image content, we demonstrate our GRM and GAM in the first three columns and our multi-stage addition manner in the 4th and 5th columns. We can observe that worse detail preservation occurs in both reconstruction and editing in \emph{without GRM} (1st column) due to the lack of exact details extracted by GRM. Edited results own heavy artifacts in \emph{without GAM} (2nd column), such as the hat of the man and the mouth of the woman, as there is no details refinement conducted by GAM in editing. Using both modules has the best result (6th column), which demonstrates that our GRM effectively supplements details and GAM correctly suppresses artifacts. Moreover, adding residual features in a single \emph{low-level stage} (i.e., coarser feature map) leads to poor reconstruction quality (4th column), and only adding in a single \emph{high-level stage} (i.e., finer feature map) results in editing artifacts (5th column). Our multi-stage method can achieve an excellent distortion-editability trade-off.

\section{Conclusions}
In StyleGAN inversion and editing, we propose to gradually add details information for the first time, which achieves a unity of both high-quality detail preservation and high editability. In particular, a novel dual-stream framework is proposed to calculate residual features step by step and then align them with edited images. Further, We utilize a self-supervised training strategy to train both streams simultaneously. Extensive experiments have shown the effectiveness of our framework and the improvement over existing methods in terms of reconstruction and editing.

\section*{Acknowledgments}
This work is supported by the National Natural Science Foundation of China under Grant U19A2057 and the National Science Fund for Excellent Young Scholars under Grant 62222212.

\bibliography{aaai24}
\clearpage

\appendix
\section{Appendix}

We provide more details of training (section A.1), the networks of our framework (section A.2), and the multi-stage manner (section A.3). Then we show more experimental results (section A.4). Further, we analyze the limitations of our method and some future directions (section A.5).

\subsection{Training Details}
We employ the pre-trained generator from StyleGAN2 \cite{karras2020analyzing}, and the pre-trained encoder from e4e \cite{tov2021designing} in all our experiments. Our framework is trained on RTX3090 GPUs with batch size 8 on $1024 \times 1024$ resolution (face domain) or $384 \times 512 $ (car domain). We fix the learning rate of our framework at 0.0001, and the learning rate of discriminator $D$ is also 0.0001. The number of training iterations is 70000. $\lambda_{lpips}$ is 0.8, $\lambda_{adv}$ is set to 0.001, $\lambda_{aligner}$ is 0.1, $\lambda_{f}$ is set to 0.01, and $\lambda_{id}$ is set to 0.1 for face domain and 0.5 for car domain.

\begin{figure}[h]
\centering
\includegraphics[width=\columnwidth]{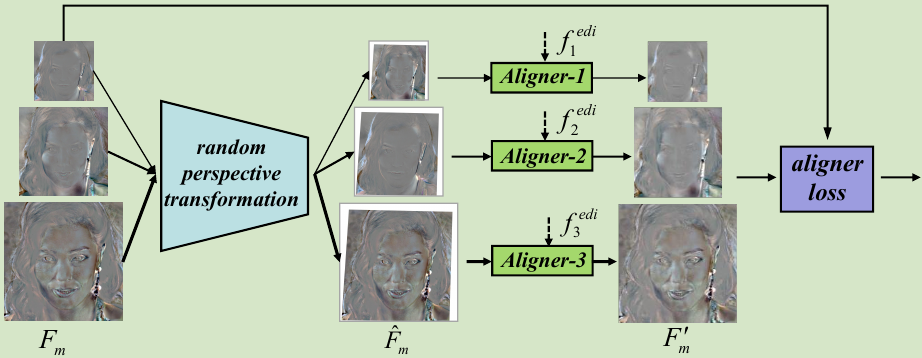} 
\caption{Illustration of our self-supervised training strategy.}
\label{fig9}
\end{figure}

Moreover, we illustrate our self-supervised training strategy in Fig.\ref{fig9}. Inspired by HFGI \cite{wang2022high}, we train the editing ability of our framework without labeled edited images by constructing the misaligned feature pairs for each $GAM$. The distortion scale of the \emph{random perspective transformation} is set to 0.15. In this figure, we provide an image-level visualization, while the processes are actually conducted at the feature level. What's more, we find that utilizing a progressive training strategy to train our framework, which means we first train the $GRM$ and $GAM$ of one stage and then train the $GRM$ and $GAM$ of the next stage with the previous stages fixed, can achieve similar results but consume fewer resources than training together. 

\subsection{Networks Details}
\begin{figure}[t]
\centering
\includegraphics[width=0.8\columnwidth]{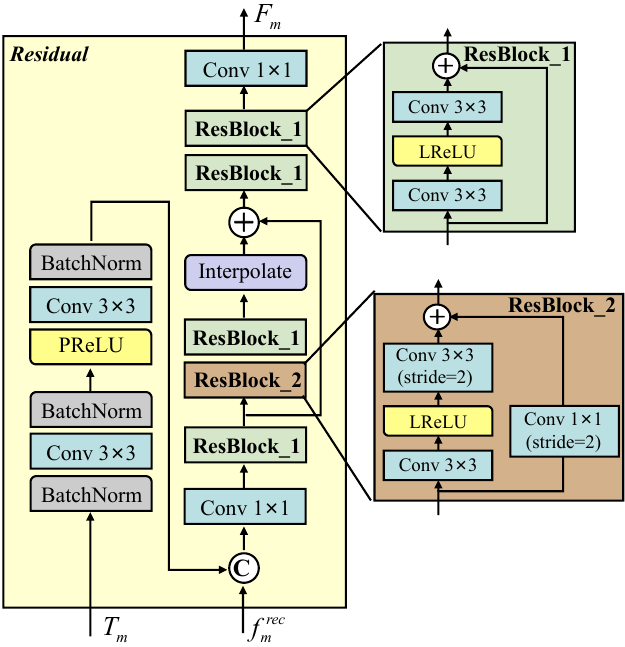} 
\caption{Detailed structure of our $Residual$ block. }
\label{fig10}
\end{figure}

\begin{figure}[t]
\centering
\includegraphics[width=0.45\columnwidth]{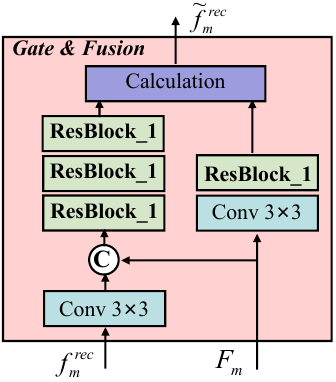} 
\caption{Detailed structure of our $Gate \& Fusion$ block.}
\label{fig11}
\end{figure}

We illustrate the details of our networks in Fig.\ref{fig10} and Fig.\ref{fig11}. Our $Residual$ in Fig.\ref{fig10} consists of the Convolution layer, $\mathrm{PReLU}$ layer, $\mathrm{BatchNorm}$ layer, and designed $\mathrm{ResBlock}$ layer from \cite{pehlivan2023styleres}, especially, $\mathrm{ResBlock\_2}$ layer can reduce the resolution to half, and $\mathrm{Interpolate}$ doubles the resolution by using linear interpolation. $Gate\&Fusion$ in Fig.\ref{fig11} is combined with several $\mathrm{ResBlock\_1}$ layers, where $\mathrm{LReLU}$ stands for Leaky ReLU \cite{xu2020reluplex}. As we insert our $GRM$ and $GAM$ between every two stages, there is almost no scale change in the input and output of our modules. Moreover, $Aligner$ has been shown in our main paper and $Gate\&Fusion$ block of our $GAM$ is the same as the block of $GRM$. During inference, our framework can bring a huge improvement in performance, but will only slightly increase the time cost, because our proposed GAM and GRM only conduct fast feedforward calculations along with the generation process.

\begin{figure*}[t]
\centering
\includegraphics[width=\textwidth]{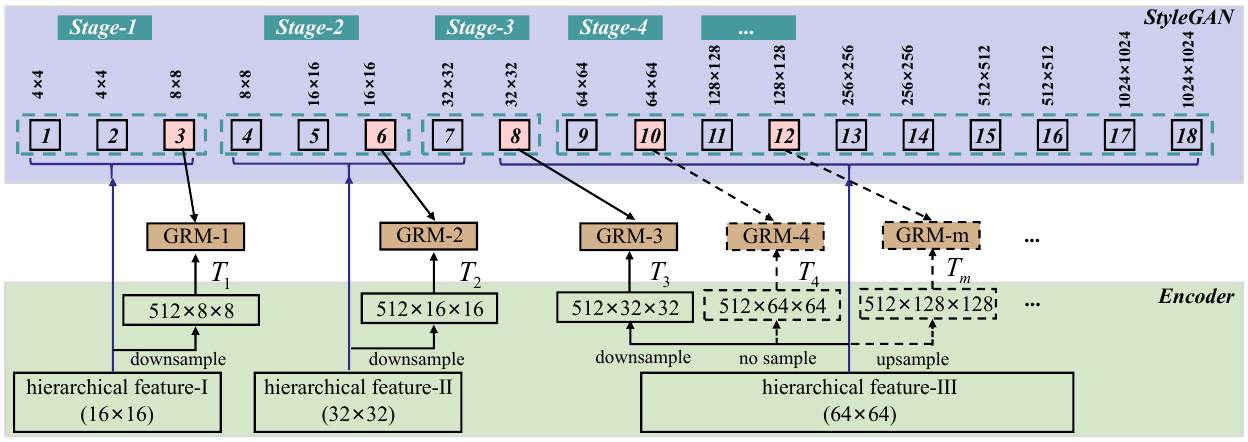} 
\caption{Detailed structure of our encoder and generator. It shows how the generator blocks and latent codes are grouped into multiple stages and how the hierarchical features correspond to each stage.}
\label{fig12}
\end{figure*}

\setlength{\tabcolsep}{9pt}
\begin{table*}[t]
\begin{center}
\renewcommand{\arraystretch}{1.08}
\begin{tabular}{cccccc}
\hline
\multicolumn{1}{c}{Refined Blocks} & \multicolumn{1}{c}{ID($\uparrow$)} & \multicolumn{1}{c}{L2($\downarrow$)} & \multicolumn{1}{c}{LPIPS($\downarrow$)} & \multicolumn{1}{c}{ID(for pose, $\uparrow$)} & \multicolumn{1}{c}{Feature Scales} \\ \hline
    no   & 0.499 & 0.053 & 0.394 & 0.464 & - \\
    3rd   & 0.504 & 0.048 & 0.379 & 0.470 & $8\times8$ \\
    3rd, 6th   & 0.568 & 0.034 & 0.332 & 0.518 & $8\times8, 16\times16$ \\ 
    \textbf{3rd, 6th, 8th}   & 0.813 & 0.015 & 0.269 & 0.677 & \textbf{$8\times8, 16\times16, 32\times 32$} \\
    3rd, 6th, 8th, 10th   & 0.873 & 0.011 & 0.246 & 0.636 & $8\times8, 16\times16, 32\times 32, 64\times 64$ \\
    3rd, 4th, 5th, 6th, 7th, 8th   & 0.816 & 0.015 & 0.265 & 0.665 & $(8\times8, 16\times16, 32\times32)\times2$ \\ \hline
\end{tabular}
\caption{Quantitative comparisons of different refined blocks, $\downarrow$ indicates lower is better while $\uparrow$ indicates higher is better.}
\label{table3}
\end{center}
\end{table*}

\subsection{Stages Details}
We illustrate the details of our encoder and generator in Fig.\ref{fig12}. To be specific, for $1024 \times 1024$ resolution in StyleGAN2, there are $N=18$ blocks and latent codes in total, pre-trained encoder from e4e \cite{tov2021designing} use a pyramid structure to extra the hierarchical features. These hierarchical features are utilized to generate corresponding latent codes, for example, the smallest feature ($16\times16\times512$) generates the 1st, 2nd, and 3rd latent codes, the middle feature ($32\times32\times512$) generates the 4th, 5th, 6th and 7th latent codes, and the largest feature ($64\times64\times512$) generates latent codes from the 8th to the 18th. 

As strong correlations already exist between latent codes and hierarchical features, we refine a block by only using the corresponding hierarchical feature. If the scale of current hierarchical features cannot match the scale of current feature maps from the generator, we will utilize an upsampling or downsampling operation to change the scale, for example, we downsample the smallest hierarchical feature ($16\times16\times512$) to $T_1$ ($8\times8\times512$) for matching the scale of the 3rd block ($8\times8\times512$). 

Considering the tradeoff of effectiveness and resource requirements, we mainly set our framework to a 4-stage process where the 3rd, 6th, and 8th blocks of the StyleGAN generator are refined, we ensure each hierarchical feature is employed to refine a block. In this way, we can achieve not only satisfactory results but also less resource consumption. An extended ablation study of how to choose blocks in our multi-stage process is conducted in Tab.\ref{table3}. We compare different choices in this table, from the 1st row to the 6th row of the table, it includes \emph{one stage} only (no), \emph{two stages} (3rd), \emph{three stages} (3rd, 6th), \emph{four stages} (3rd, 6th, 8th), \emph{five stages} (3rd, 6th, 8th, 10th), and \emph{dense stages} (refine blocks from 3rd to 8th). We ignore the low resolution (1st and 2nd blocks) and higher resolution (from 11th to 18th blocks), as the low-resolution feature map is too small to conduct valuable calculations while the high-resolution needs too many resources \cite{karras2019style}. We report the ID, L2, and LPIPS to evaluate reconstruction quality and the ID (pose) to evaluate editing. We can observe that with the increment of feature scales, the results gradually become better in both reconstruction and editing, but the resource consumption actually becomes larger and larger. The \emph{dense stages} achieves similar improvement with the \emph{three stages} but needs more calculation resources. Moreover, the \emph{three stages} owns the best editing ID score.

\subsection{More Experimental Results}
\begin{figure*}[t]
\centering
\includegraphics[width=\textwidth]{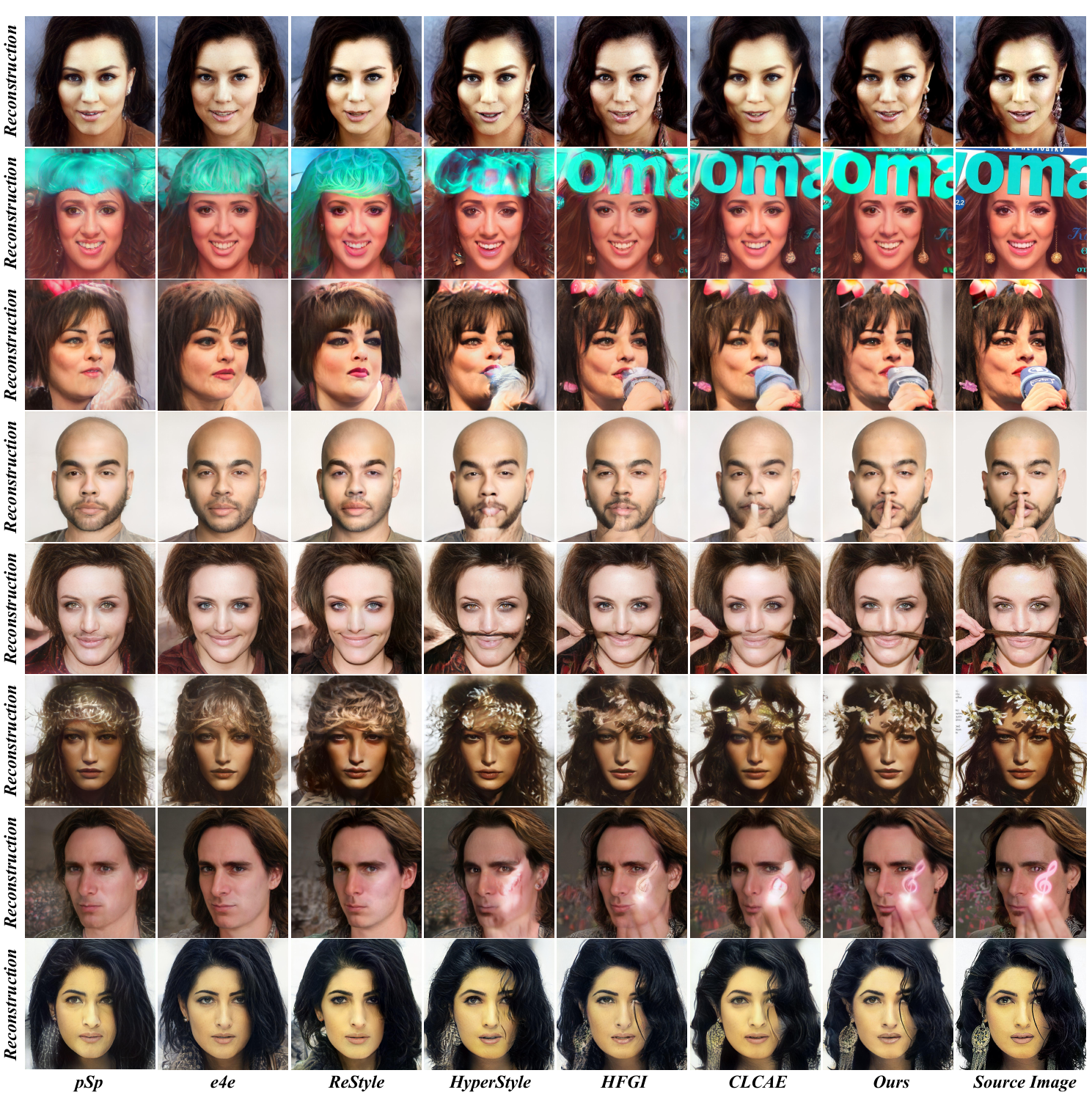} 
\caption{Visual comparisons on image reconstruction.}
\label{fig13}
\end{figure*}

\begin{figure*}[t]
\centering
\includegraphics[width=\textwidth]{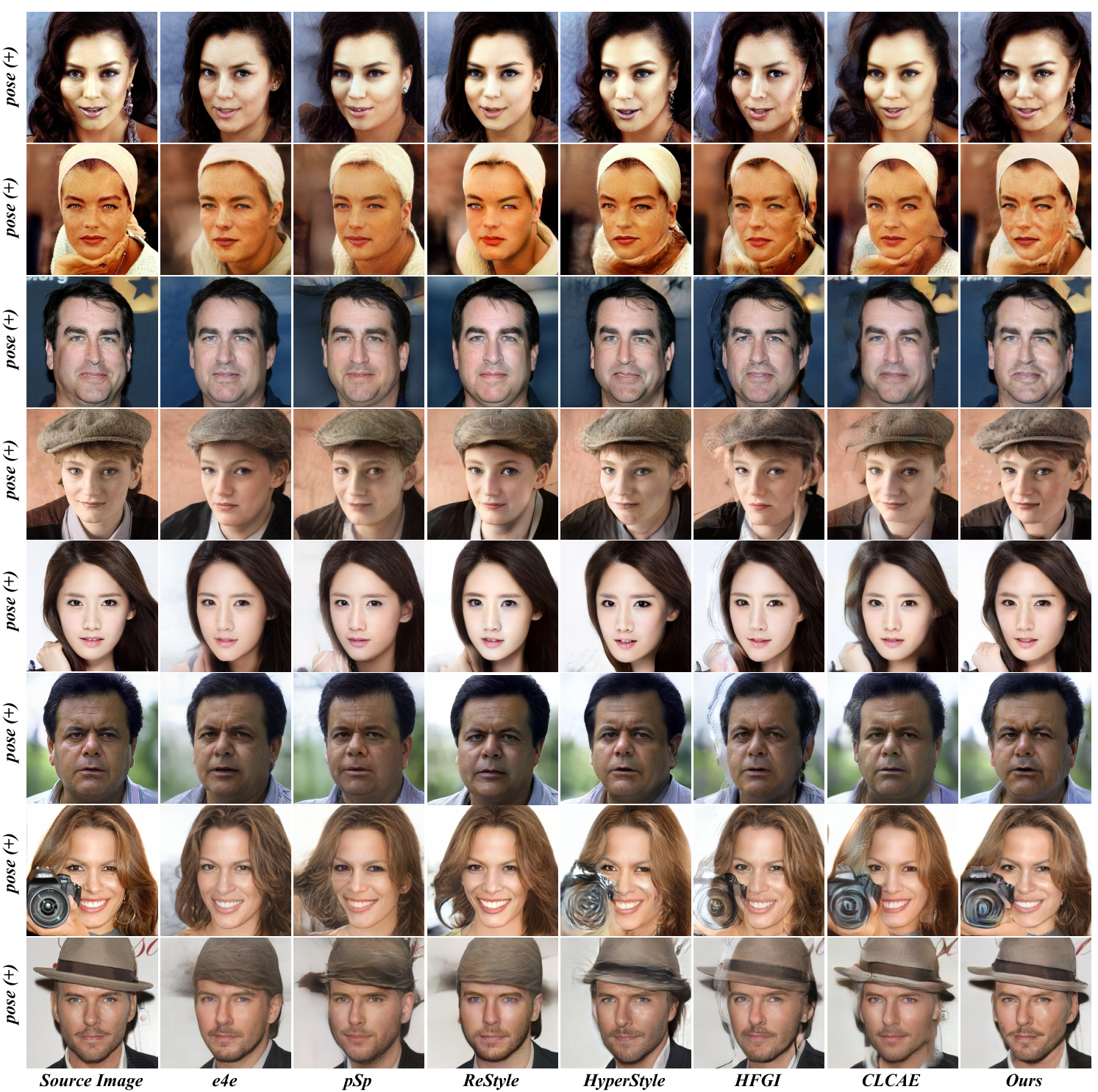} 
\caption{Visual comparisons on image editing (pose).}
\label{fig14}
\end{figure*}

\begin{figure*}[t]
\centering
\includegraphics[width=\textwidth]{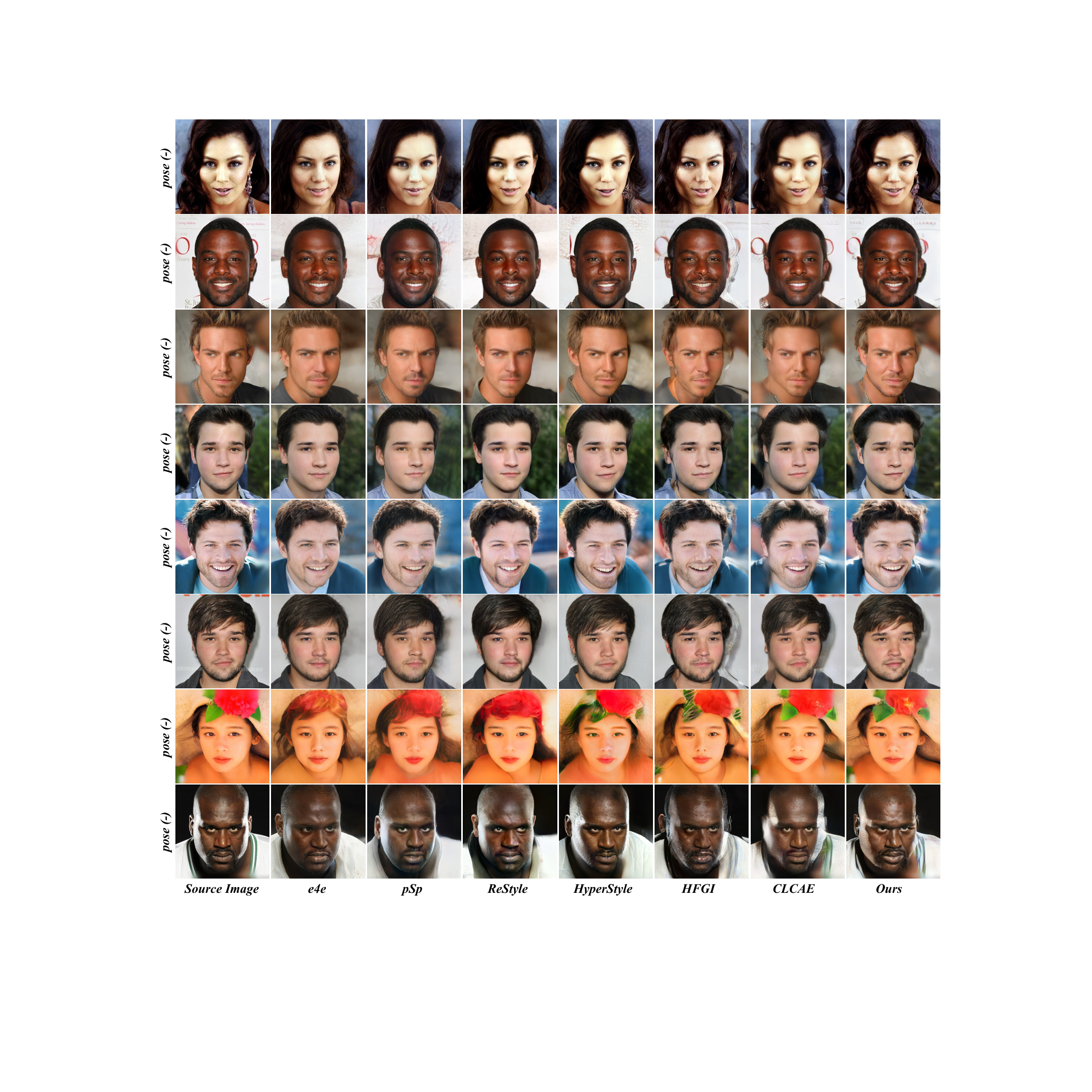} 
\caption{Visual comparisons on image editing (pose).}
\label{fig15}
\end{figure*}

\begin{figure*}[t]
\centering
\includegraphics[width=\textwidth]{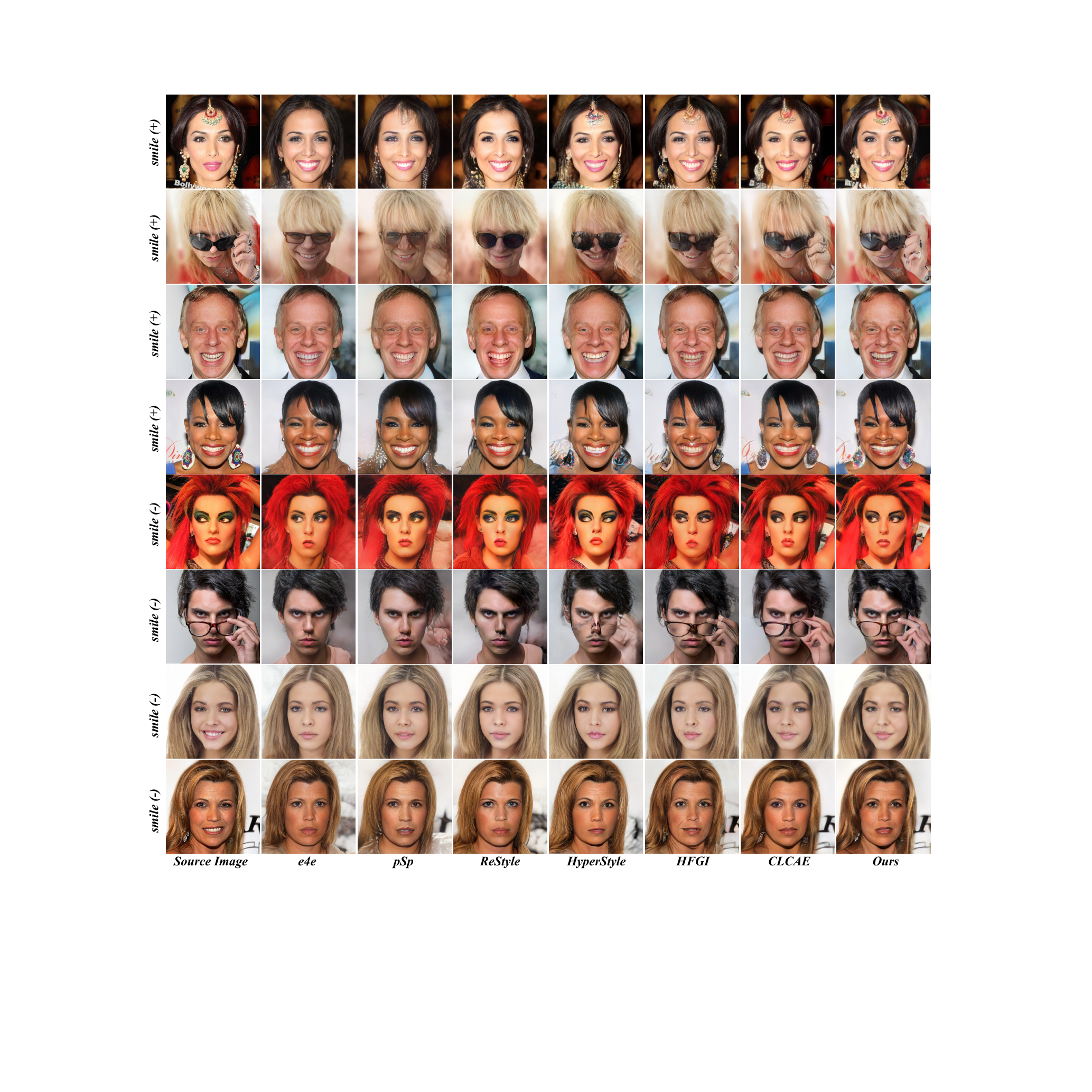} 
\caption{Visual comparisons on image editing (smile).}
\label{fig16}
\end{figure*}

\begin{figure*}[t]
\centering
\includegraphics[width=\textwidth]{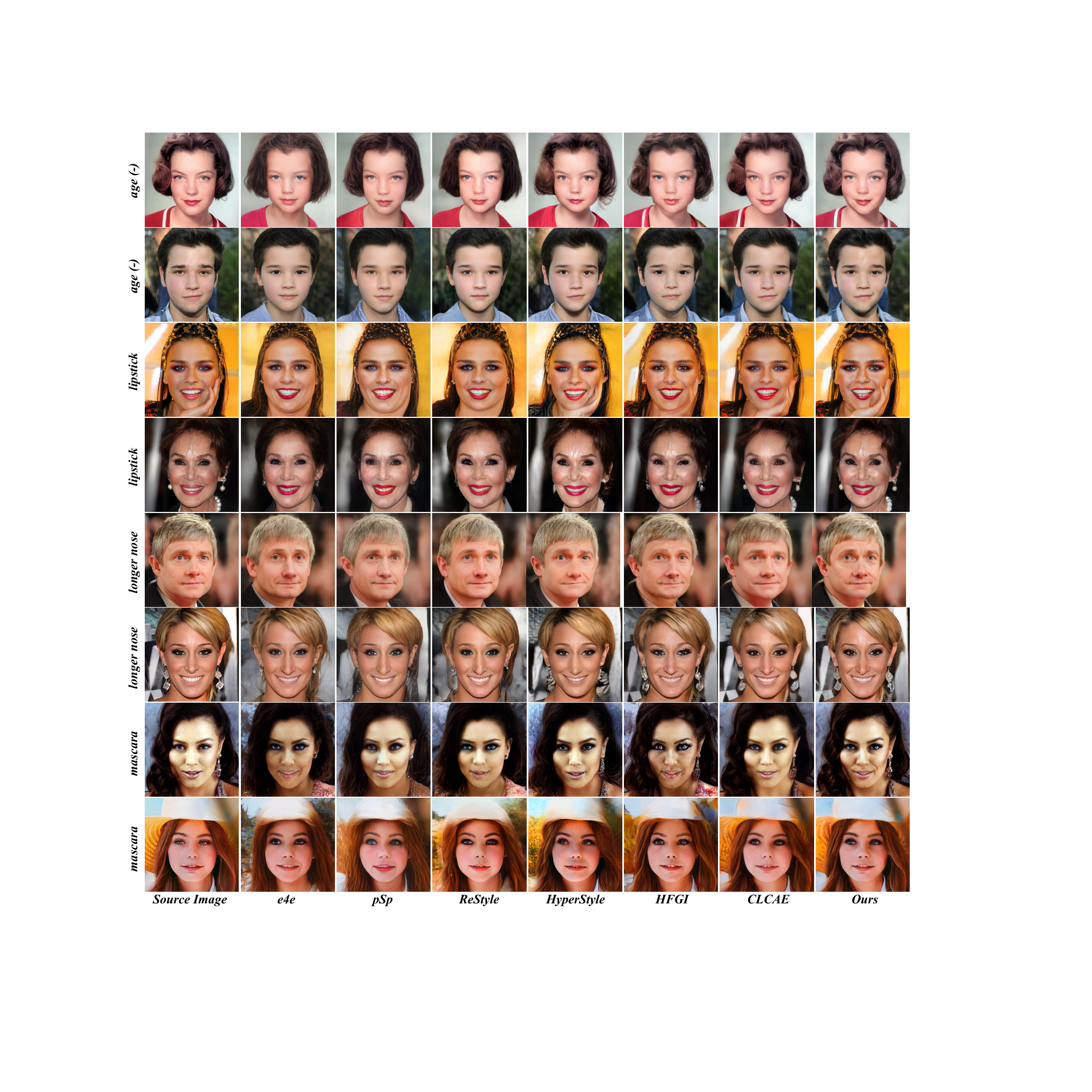} 
\caption{Visual comparisons on various image editing scenes.}
\label{fig17}
\end{figure*}

\begin{figure*}[t]
\centering
\includegraphics[width=\textwidth]{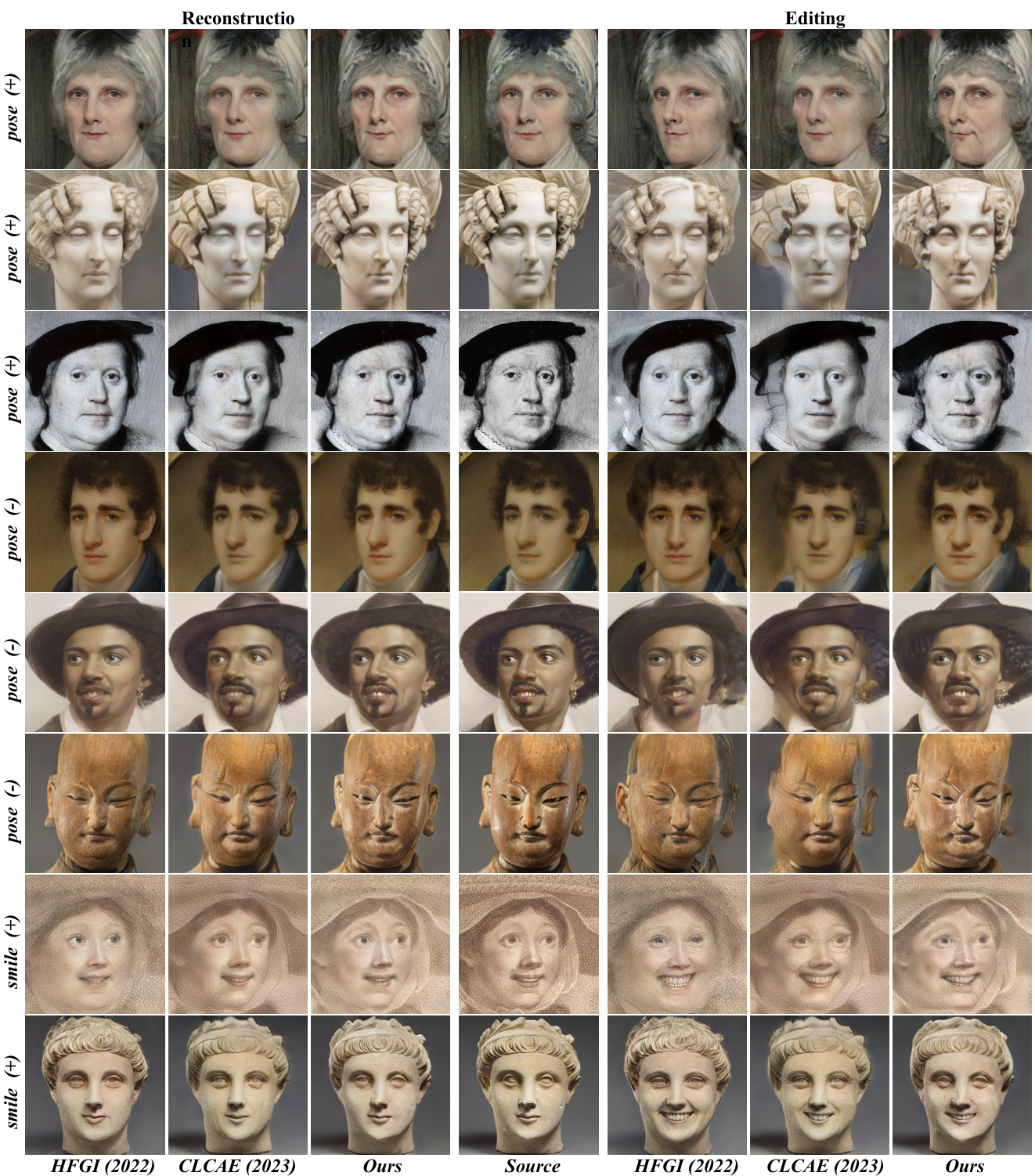} 
\caption{Additional results of the artistic portrait domain on both image reconstruction and editing.}
\label{fig18}
\end{figure*}

In our experiments, both qualitative and quantitative results are obtained by employing the released codes and checkpoints of these state-of-the-art methods in the same environment. We provide some quantitative comparisons of ablation studies in Tab.\ref{table4} and more visualization results of qualitative experiments from Fig.\ref{fig13} to Fig.\ref{fig18}, where pSp \cite{richardson2021encoding}, e4e \cite{tov2021designing}, ReStyle \cite{alaluf2021restyle}, HyperStyle \cite{alaluf2022hyperstyle}, HFGI \cite{wang2022high}, CLCAE \cite{liu2023delving} and our method are compared. 

We report the ID and L2 to evaluate reconstruction quality and also report the ID (pose) to evaluate editing in Tab.\ref{table4}. In addition to our previous ablation studies on GAM/GRM and multi-stage manner, we further show the role of hierarchical features (from the encoder) by only utilizing different single-level features from one of the hierarchical encoder layers (from I to III). These features belong to different granularity, our model can better keep the image content consistent by employing corresponding different-level encoder features in different generation stages. All results demonstrate the effectiveness of our designs. 

Fig.\ref{fig13} shows some reconstruction examples. We show \emph{pose(+)} editing (i.e., heads turn left) results by using InterfaceGAN \cite{shen2020interfacegan} with magnitude 2 in Fig.\ref{fig14}. In Fig.\ref{fig15}, \emph{pose(-)} editing (i.e., heads turn right) results are shown. Fig.\ref{fig16} shows \emph{smile(+)} and \emph{smile(-)} editing results. Fig.\ref{fig17} illustrates several editing scenes, including \emph{lipstick}, \emph{nose} and \emph{mascara} edited by GANspace \cite{harkonen2020ganspace}, and \emph{age(-)} edited by InterfaceGAN. Some additional results of the artistic portrait domain are demonstrated in Fig.\ref{fig18}.

\setlength{\tabcolsep}{4pt}
\begin{table}[t]
\begin{center}
\renewcommand{\arraystretch}{1.08}
\begin{tabular}{c|ccc}
    \hline
                           & ID($\uparrow$)             & L2($\downarrow$)             & ID(editing)   \\ \hline
    w/o GAM                & 0.708          & 0.021          & 0.496          \\
    w/o GRM                & 0.583          & 0.032          & 0.543          \\
    w/o GAM\&GRM           & 0.509          & 0.042          & 0.484          \\ \hline
    one stage(low)         & 0.504          & 0.048          & 0.470          \\
    one stage(high)        & 0.763          & 0.019          & 0.605          \\ \hline
    only hierarchical I    & 0.702          & 0.023          & 0.614          \\
    only hierarchical II   & 0.742          & 0.020          & 0.624          \\
    only hierarchical III  & 0.785          & 0.018          & 0.615          \\ \hline
    full (ours)            & \textbf{0.813} & \textbf{0.015} & \textbf{0.677} \\ \hline
    \end{tabular}
\caption{Quantitative comparisons of ablation studies.}
\label{table4}
\end{center}
\end{table}

\subsection{Future Directions}
We utilized a self-supervised training strategy to train our dual-stream framework, and we have achieved promising reconstruction and editing results. However, there are still some difficult cases for us, such as wrinkles addition, and pale hair editing (from black to white). One possible reason is that random transformations in the self-supervised strategy give our framework the ability to handle various editing scenes, but the addition of facial details elements and the modification of color cannot be effectively simulated, making them difficult to edit. We will explore using some other simulation techniques to simulate facial details addition and color modifications, we can also design a supervised training strategy by utilizing some labeled edited images or synthesized images to solve these difficult cases.

\end{document}